\documentclass{llncs}

\usepackage[utf8]{inputenc}
\usepackage{rotating}
\usepackage{grffile}
\usepackage{subfigure}
\usepackage{amsmath}
\usepackage{amssymb}

\begin{document}

\title { The Link Prediction Problem \\ in Bipartite Networks }

\author{Jérôme Kunegis \and Ernesto W. De Luca \and Sahin Albayrak}

\institute{
  DAI Lab, Technische Universität Berlin \\
  Ernst-Reuter-Platz 7 \\
  D-10587 Berlin, Germany \\
  \email{\{jerome.kunegis,ernesto.deluca,sahin.albayrak\}@dai-labor.de}
}

\maketitle

\begin{abstract}
We define and study the link prediction problem in bipartite networks,
specializing general link prediction algorithms to the bipartite case.
In a graph, a link prediction function of two vertices denotes the
similarity or proximity of the vertices.  Common link prediction
functions for general graphs are defined using paths of length two
between two nodes.  Since in a bipartite graph adjacency vertices can
only be connected by paths of odd lengths, these functions do not apply
to bipartite graphs.  Instead, a certain class of graph
kernels (spectral transformation kernels) can be generalized to
bipartite graphs when the positive-semidefinite kernel constraint is
relaxed.  This generalization is realized by the odd component of the
underlying spectral transformation.  This construction leads to several
new link prediction pseudokernels such as the matrix hyperbolic sine,
which we examine for rating graphs, authorship graphs, folksonomies,
document--feature networks and other types of bipartite networks.
\end{abstract}

\section{Introduction}
In networks where edges appear over time, the problem of predicting such
edges is called \emph{link prediction}~\cite{b256,b530}.  Common
approaches to link prediction can be described as \emph{local} when only
the immediate neighborhood of vertices is considered and \emph{latent}
when a latent model of the network is used.  An example for local link
prediction methods is the triangle closing model, and these models are
conceptually very simple.  Latent link prediction methods are
instead derived using algebraic graph theory: The network's adjacency
matrix is decomposed and a transformation is applied to the network's
spectrum. This approach is predicted by several graph growth models and
results in \emph{graph kernels}, positive-semidefinite functions of the
adjacency matrix~\cite{b549}.

Many networks contain edges between two types of entities, for instance
item rating graphs, authorship graphs and document--feature networks.
These graphs are called bipartite~\cite{b531}, and while they are a
special case of general graphs, link prediction methods cannot be
generalized to them.  As we show in
Section~\ref{sec:bipartite-link-prediction}, this is the case for all
link prediction functions 
based on the triangle closing model, as well as all
positive-semidefinite graph kernels.  Instead, we will see that their
odd components can be used, in Section~\ref{sec:algebraic}.  For each positive-semidefinite graph
kernel, we derive the corresponding odd pseudokernel.  One example is
the exponential graph kernel $\exp(\lambda)$.  Its odd component is
$\sinh(\lambda)$, the hyperbolic sine.  We also introduce the bipartite
von~Neumann pseudokernel, and study the bipartite versions of
polynomials with only odd powers.  We show experimentally (in
Section~\ref{sec:experiments}) how these odd
pseudokernels perform on the task of link prediction in bipartite
networks in comparison to their positive counterparts, and give an
overview of their relative performances .  We also sketch their usage for 
detecting near-bipartite graphs.


\section{Bipartite Link Prediction}
\label{sec:bipartite-link-prediction}
The link prediction problem is usually defined on unipartite graphs, where
common link prediction algorithms make several assumptions~\cite{b461}:  
\begin{itemize}
\item \emph{Triangle closing}:  New edges tend to form triangles. 
\item \emph{Clustering}: Nodes tend to form well-connected clusters in
  the graph.   
\end{itemize}
In bipartite graphs these assumptions are not true, since triangles and
larger cliques cannot appear.  Other assumptions have therefore to be
used.  While a unipartite link prediction algorithm technically applies
to bipartite graphs, it will not perform well.  Methods based on common
neighbors of two vertices will for instance not be able to predict
anything in bipartite graphs, since two vertices that would be connected
(from different clusters) do not have any common neighbors.

Several important classes of networks are bipartite: authorship
networks, interaction networks, usage logs, ontologies and many more.
Many unipartite networks (such as coauthorship networks) can be
reinterpreted as bipartite networks when edges or cliques are modeled as
vertices.  In these cases, special bipartite link prediction
algorithms are necessary.  The following two sections will review local
and algebraic link prediction methods for bipartite graphs.
Examples of specific networks of these types will be given in
Section~\ref{sec:experiments}.   

\begin{figure}[t]
  \centering
  \subfigure[Unipartite network]{\includegraphics[width=.40\textwidth]{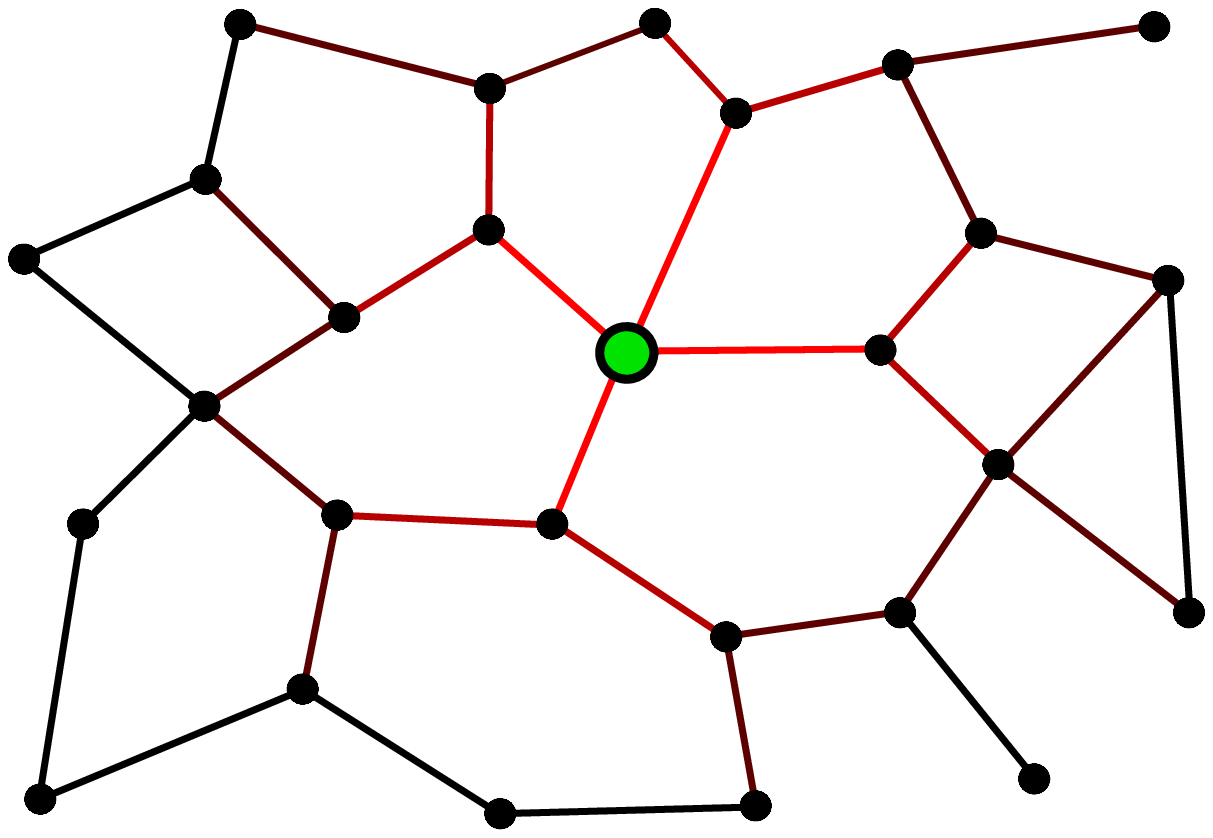}\,}
  \subfigure[Bipartite network]{\includegraphics[width=.52\textwidth]{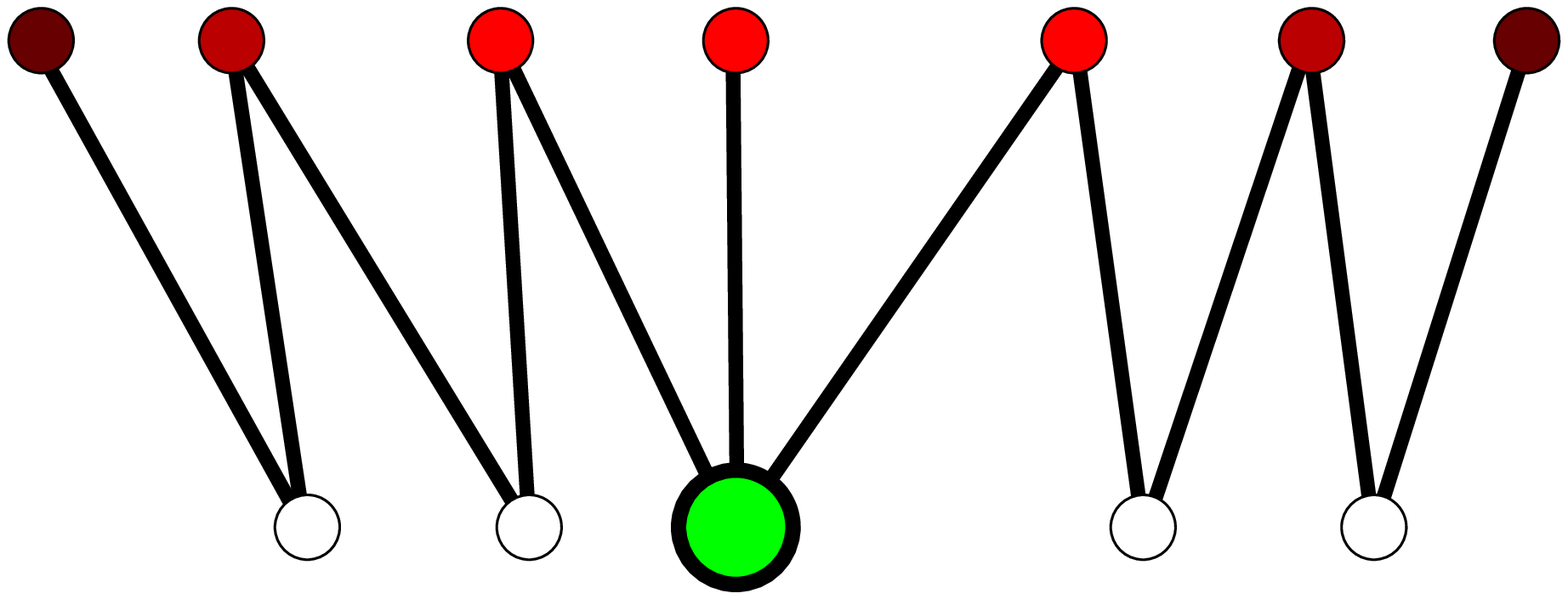}}
  \caption{
    Link prediction by spreading activation in unipartite and bipartite
    networks.  In the unipartite case, all paths are used.  In the
    bipartite case, only paths of odd length need to be considered.  In
    both cases, the weight of paths is weighted in inverse proportion to
    path length.
  }
  \label{fig:unibip}
\end{figure}

\subsubsection{Definitions}
Given an undirected graph $G=(V,E)$ with vertex set $V$ and edge set $E$,
its adjacency matrix $A\in \mathbb R^{V\times V}$ is defined as
$A_{uv}=1$ when $(u,v)\in E$ and $A_{uv}=0$ otherwise.  
For a bipartite graph $G=(V + W, E)$, the adjacency matrix can be
written as $A= \left[ 0 \, B; B^T \, 0 \right] $, where $B \in \mathbb
R^{V \times W}$ is the biadjacency matrix of $G$. 

\subsection{Local Link Prediction Functions}
Some link prediction functions only depend on the immediate neighborhood
of two nodes; we will call these functions local link prediction
functions~\cite{b256}.  

Let $u$ and $v$ be two nodes in the graph for which a link prediction
score is to be computed.  Local link prediction functions depend on the
common neighbors of $u$ and $v$.  In the bipartite link prediction
problem, $u$ and $v$ are in different clusters, and thus have no common
neighbors.  The following link prediction functions are therefore not
applicable to bipartite graphs: The number of common
neighbors~\cite{b256}, the measure of Adamic and Adar~\cite{b475} and
the Jaccard coefficient~\cite{b256}.
These methods are all based on the \emph{triangle closing} model, which
is not valid for bipartite graphs. 

\subsubsection{Preferential Attachment}
Taking only the degree of $u$ and $v$ into account for link prediction
leads to the \emph{preferential attachment} model~\cite{b439}, which can
be used as a model for more complex methods such as modularity
kernels~\cite{b401,b413}.

If $d(u)$ is the number of neighbors of node $u$, the preferential
attachment models gives a prediction between $u$ and $v$ of $d(u)
d(v)/(2|E|)$.  The factor $1/(2|E|)$
normalizes the sum of predictions for a vertex to its degree. 

\section{Algebraic Link Prediction Functions}
\label{sec:algebraic}
Link prediction algorithms that not only take into account the immediate
neighborhood of two nodes but the complete graph can be formulated using
algebraic graph theory, whereby a decomposition of the graph's adjacency
matrix is computed~\cite{b285}.  By considering transformations of a
graph's adjacency matrix, link prediction methods can be defined and
learned.  Algebraic link prediction methods are motivated by their
scalability and their learnability.  They are scalable because they rely
on a model that is built once and which makes computation of
recommendations fast.  These models correspond to decomposed matrices
and can usually be updated using iterative algorithms~\cite{b371}.  In
contrast, local link prediction algorithms are \emph{memory-based},
meaning they access the adjacency data directly during link prediction.
Algebraic link prediction methods are learnable because their parameters
can be learned in a unified way~\cite{kunegis:spectral-transformation}.

In this section, we describe how algebraic link prediction methods apply
to bipartite networks.  Let $G=(V,E)$ be a (not necessarily bipartite)
graph.  Algebraic link prediction algorithms are based on the eigenvalue
decomposition of its adjacency matrix $A$:
\begin{eqnarray*}
 A &=& U \Lambda U^T 
\end{eqnarray*}
To predict links, a \emph{spectral transformation} is usually applied:
\begin{eqnarray*}
F(A) &=& U F(\Lambda) U^T
\end{eqnarray*}
where $F(\Lambda)$ applies a real function $f(\lambda)$ to each eigenvalue
$\lambda_i$.  $F(A)$ then contains link prediction scores that, for each
node, give a ranking of all other nodes, which is then used for link
prediction. 
If $f(\lambda_i)$ is positive, $F$ is a graph kernel, otherwise, we will
call $F$ a pseudokernel. 

Several spectral transformations can be written as polynomials of the
adjacency matrix in the following way.  The matrix power $A^i$ gives,
for each vertex pair $(u,v)$, the number of paths of length $i$ between
$u$ and $v$.  Therefore, a polynomial of $A$ gives, for a pair $(u,v)$,
the sum of all paths between $u$ and $v$, weighted by the polynomial
coefficients.  This fact can be exploited to find link prediction
functions that fulfill the two following requirements:
\begin{itemize}
\item The link prediction score should be higher when two nodes are
  connected by \emph{many} paths.
\item The link prediction score should be higher when paths are
  \emph{short}.  
\end{itemize}
These requirements suggest the use of polynomials $f$ with decreasing
coefficients.  

\subsection{Odd Pseudokernels}
In bipartite networks, only paths of odd length are significant, since
an edge can only appear between two vertices if they are already
connected by paths of odd lengths.  Therefore, only odd powers are
relevant, and we can restrict the spectral transformation to odd
polynomials, i.e. polynomials with odd powers. 

The resulting spectral transformation is then an odd function and except
in the trivial and undesired case of a constant zero function, will be
negative at some point.  Therefore, all spectral transformations
described below are only pseudokernels and not kernels. 

\subsubsection{The Hyperbolic Sine}
In unipartite networks, a basic link prediction function is given by the
matrix exponential of the adjacency matrix~\cite{b137,b244,b263}.  
The matrix exponential can be derived by considering the sum 
\begin{eqnarray*}
\exp(\alpha A) &=& \sum_{i=0}^\infty \frac {\alpha^i} {i!} A^i
\end{eqnarray*}
where coefficients are decreasing
with path length.  Keeping only the odd component, we arrive at the
matrix hyperbolic sine~\cite{b479}.  
\begin{eqnarray*}
\sinh(\alpha A) &=& \sum_{i=0}^\infty \frac {\alpha^{1+2i}}  {(1+2i)!} A^{1+2i}
\end{eqnarray*}

\begin{figure}[t]
  \centering
  \includegraphics[width=.65\textwidth]{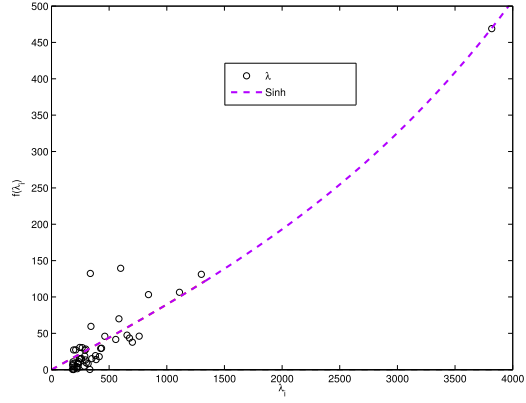}
  \caption{
    In this curve fitting plot of the Slovak Wikipedia, the hyperbolic
    sine is a good match, indicating that the hyperbolic sine
    pseudokernel performs well.   
  }
  \label{fig:sinh}
\end{figure}

Figure~\ref{fig:sinh} shows the hyperbolic sine applied to the
(positive) spectrum of the bipartite Slovak Wikipedia user--article edit
network. 

\subsubsection{The Odd von Neumann Pseudokernel}
The von Neumann kernel for unipartite graphs is given by the following
expression~\cite{b137}. 
\begin{eqnarray*}
  K_\mathrm{NEU} (A) = (I - \alpha A)^{-1} &=& \sum_{i=0}^\infty \alpha^i A^i 
\end{eqnarray*}
We call its odd component the odd von Neumann pseudokernel:
\begin{eqnarray*}
  K_\mathrm{NEU}^\mathrm{odd} (A) = \alpha A (I - \alpha^2 A^2)^{-1} &=&
    \sum_{i=0}^\infty \alpha^{1+2i} A^{1+2i} 
\end{eqnarray*}

\begin{figure}[t]
  \centering
  \includegraphics[width=.8\textwidth]{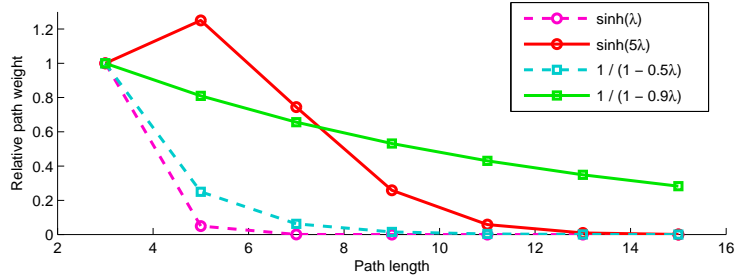}
  \caption{
    Comparison of several odd pseudokernels:  the hyperbolic sine and
    the odd von Neumann pseudokernel. 
    The relative path weight is proportional to the corresponding
    coefficient in the Taylor series expansion of the spectral
    transformation. 
  }
  \label{fig:bar}
\end{figure}

The hyperbolic sine and von Neumann pseudokernels are compared in
Figure~\ref{fig:bar}, based on the path weights they produce. 

\subsubsection{Rank Reduction}
Similarly, rank reduction of the matrix $A$ can be described as a
pseudokernel.  Let $\lambda_k$ be the eigenvalue with $k$-th largest
absolute value, then rank reduction is defined by 
\begin{eqnarray*}
f(\lambda) &=& \left\{ \begin{array}{l l} 
    \lambda \quad & \mathrm{if~} |\lambda| \geq |\lambda_k| \\
    0 & \mathrm{otherwise}
\end{array} \right. 
\end{eqnarray*}

This function is odd, but does not have an (odd) Taylor series
expansion. 

\subsection{Computing Latent Graph Models}
Bipartite graphs have adjacency matrices of the form 
\begin{eqnarray*}
A &=& \left( \begin{array}{cc} & B \\ B^T & \end{array} \right)
\end{eqnarray*}
where $B$ is the biadjacency matrix of the graph.  This form can be
exploited to reduce the eigenvalue decomposition of $A$ to the equivalent
 singular value decomposition $B = \tilde U \Sigma \tilde V$.  
\begin{eqnarray*}
A &=& 
  \left( \begin{array}{cc} U & U \\ V & -V \end{array} \right) 
  \left( \begin{array}{cc} +\Sigma & \\ & -\Sigma \end{array} \right)
  \left( \begin{array}{cc} U & U \\ V & -V \end{array} \right)^T 
\end{eqnarray*}
with $U =\tilde U/\sqrt 2$, $V = \tilde V/\sqrt 2$ and
each singular value $\sigma$ corresponds to the eigenvalue pair $\{\pm
\sigma\}$.  

\subsection{Learning Pseudokernels}
\label{subsec:curve}
\begin{figure}[t]
  \centering
  \subfigure[MovieLens 10M]{\includegraphics[width=.41\textwidth]{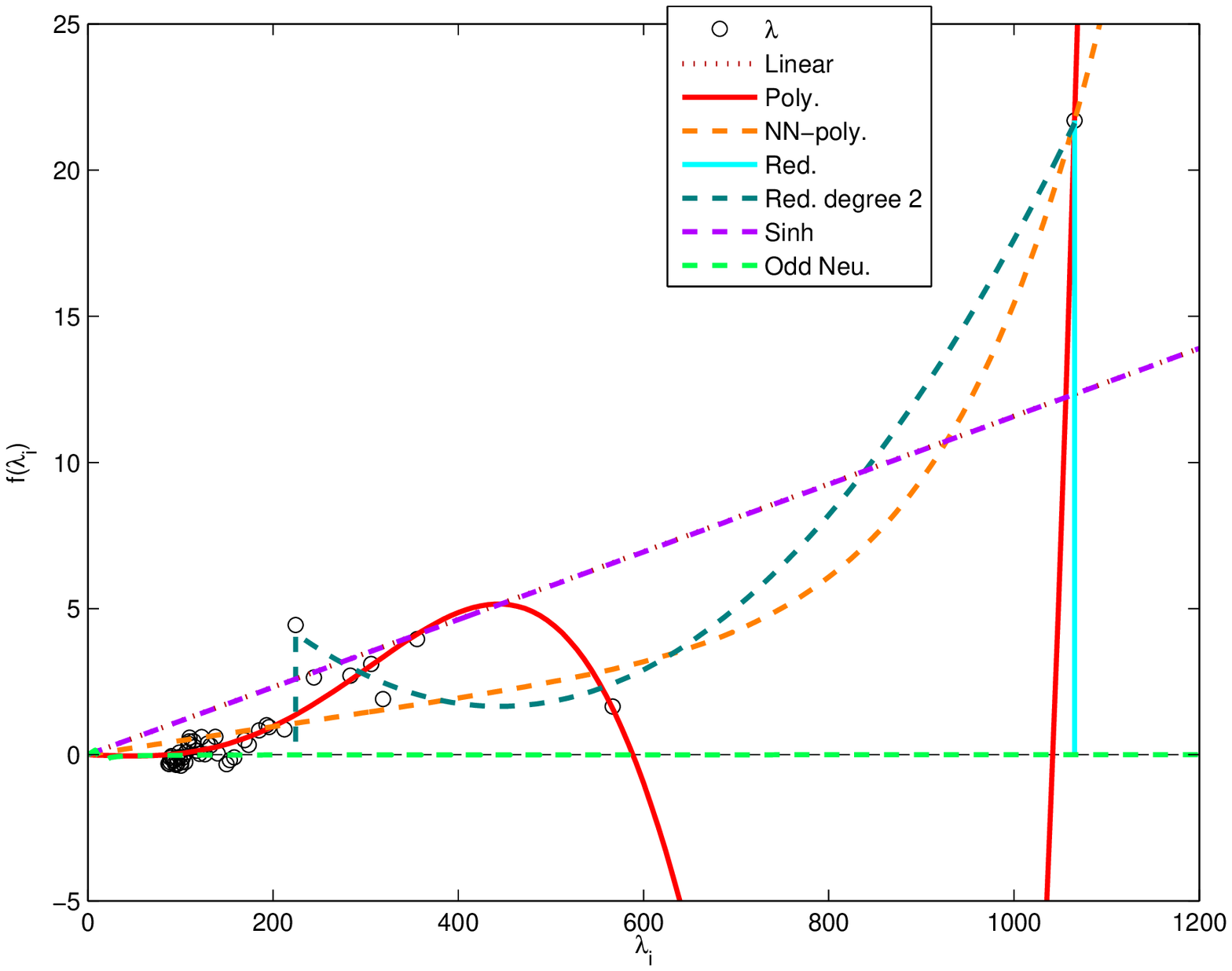}}
  \subfigure[English Wikipedia]{\includegraphics[width=.41\textwidth]{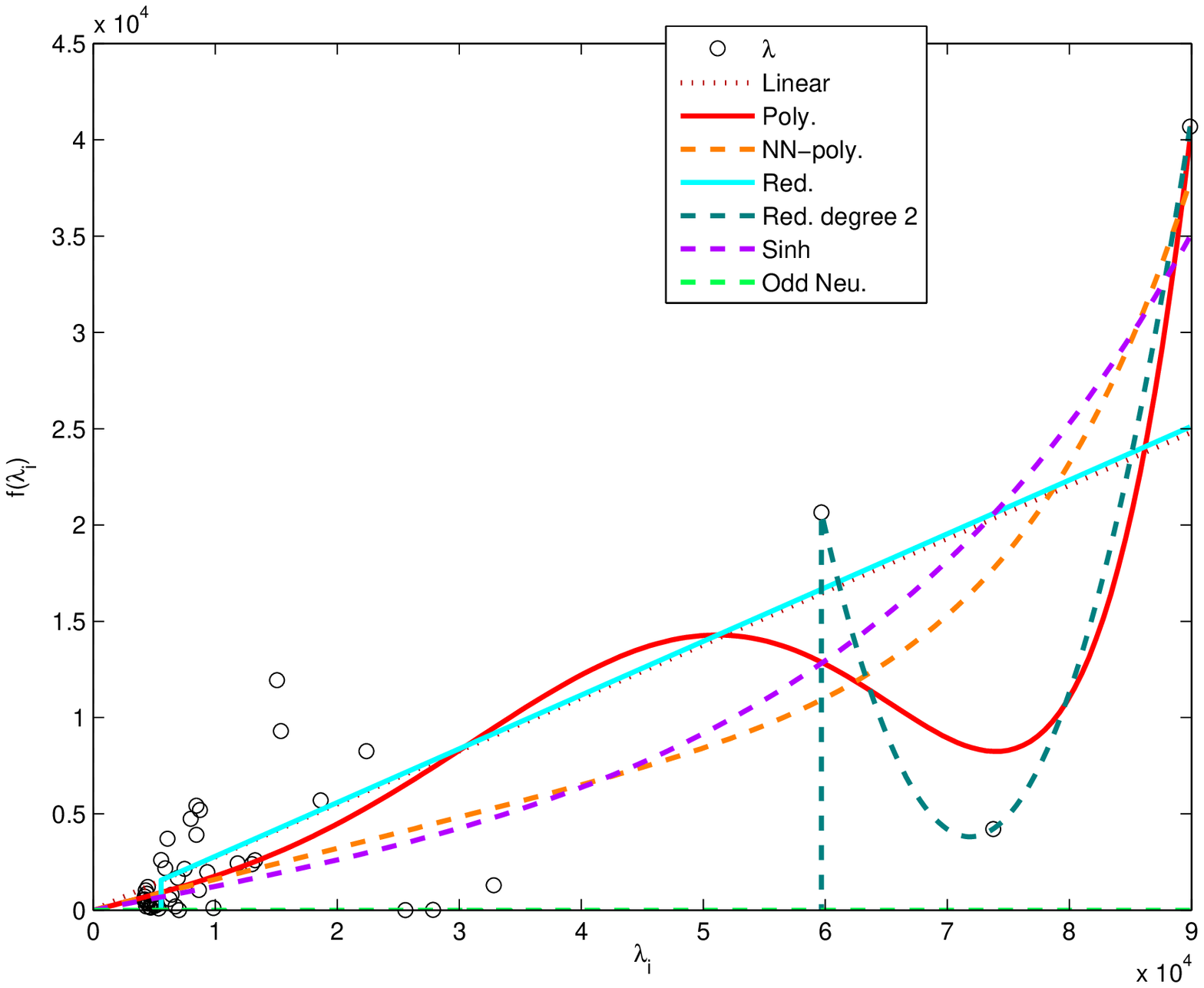}}
  \caption{
    Learning a pseudokernel that matches an observed spectral
    transformation in the MovieLens~10M rating network and English
    Wikipedia edit history. 
  }
  \label{fig:curve}
\end{figure}

The hyperbolic sine and the von Neumann pseudokernel are parametrized
by~$\alpha$, and rank reduction has the parameter $k$, or equivalently
$\lambda_k$.  These parameters can be learned by reducing the spectral
transformation problem to a one-dimensional curve fitting problem, as
described in~\cite{kunegis:spectral-transformation}.  In the bipartite
case, we can apply the curve fitting method to only the graph's singular
value, since odd spectral transformations fit the negative eigenvalue in
a similar way they fit the positive eigenvalues.  This kernel learning
method is shown in Figure~\ref{fig:curve}.

\section{Experiments}
\label{sec:experiments} 
As experiments, we show the performance of bipartite link prediction
functions on several large datasets, and present a simple method for
detecting bipartite or near-bipartite datasets. 

\subsection{Performance on Large Bipartite Networks}
We evaluate all bipartite link prediction functions on the following
bipartite network datasets. 
BibSonomy is a folksonomy of scientific publications~\cite{b346}.  
BookCrossing is a bipartite user--book interaction network~\cite{b523}. 
CiteULike is a network of tagged scientific papers~\cite{b349}. 
DBpedia is the semantic network of relations extracted from
Wikipedia, of which we study the five largest
bipartite relations~\cite{b332}. 
Epinions is the rating network from the product review site
Epinions.com~\cite{b367}. 
Jester is a user--joke network~\cite{b7}. 
MovieLens is a user--movie rating dataset, and a folksonomy of tags
attached to these movies~\cite{www.grouplens.org/node/73}. 
Netflix is the large user--item rating network associated
with the Netflix Prize~\cite{b520}.
The Wikipedia edit graphs are the bipartite user--article graphs of
edits on various language Wikipedias. 
The Wikipedia categories are represented by the bipartite
article--category network~\cite{download.wikimedia.org}. 
All datasets are bipartite and unweighted.  In rating datasets, we
only consider the presence of a rating, not the rating itself. 
Table~\ref{tab:results} gives the number of nodes and edges in each
dataset. 

In the experiments, we withhold 30\% of each network's edges as the test
set to predict.  For datasets in which edges are labeled by timestamps,
the test set consists of the newest edges.  The remaining training set is
used to compute link prediction scores using the preferential attachment
model and the pseudokernel learning methods described in the previous
sections.  For the pseudokernel learning methods, the training set is again
split into 70\%~/~30\% subsets for training.  Link prediction accuracy is
measured by the mean average precision (MAP), averaged over all users present
in the test set~\cite{b521}.  The evaluation results are summarized in
Table~\ref{tab:results}.  

\begin{table}
  \caption{
    Overview of datasets and experiment results.  See the text for a
    description of the datasets and link prediction methods. 
    Link prediction methods: 
    Poly:  odd polynomials, NN-poly:  odd nonnegative polynomials, Sinh:
    hyperbolic sine, Red:  rank reduction, Odd Neu:  odd von Neumann
    pseudokernel, Pref:  preferential attachment. 
  }
  \centering
  \scriptsize{
  \begin{tabular}{| l || r|r || l|l|l|l|l || l |}
    \hline
     Dataset                        & Nodes & Edges & Poly. & NN-poly. & Sinh  & Red.  & Odd Neu. & Pref. \\
\hline
\hline
BibSonomy tag-item               & 975,963 & 2,555,080 & 0.921 & \bf{0.925} & \bf{0.925} & 0.782 & 0.917 & 0.924 \\
BibSonomy user-item              & 777,084 & 2,555,080 & 0.748 & 0.771 & 0.771 & 0.645 & 0.750 & \bf{0.821} \\
BibSonomy user-tag               & 210,467 & 2,555,080 & 0.801 & 0.820 & 0.820 & 0.777 & 0.295 & \bf{0.878} \\
CiteULike tag-item               & 885,046 & 2,411,819 & 0.593 & 0.608 & 0.608 & 0.510 & 0.635 & \bf{0.698} \\
CiteULike user-item              & 754,484 & 2,411,819 & 0.853 & \bf{0.856} & \bf{0.856} & 0.735 & 0.855 & 0.838 \\
CiteULike user-tag               & 175,992 & 2,411,819 & 0.812 & 0.836 & 0.836 & 0.782 & 0.202 & \bf{0.881} \\
DBpedia artist-genre             & 47,293 & 94,861 & 0.824 & \bf{0.971} & 0.833 & 0.736 & 0.841 & 0.961 \\
DBpedia birthplace               & 191,652 & 273,695 & 0.952 & 0.977 & \bf{0.978} & 0.733 & 0.813 & 0.968 \\
DBpedia football club            & 41,846 & 131,084 & \bf{0.685} & 0.678 & 0.674 & 0.505 & 0.159 & 0.680 \\
DBpedia starring                 & 83,252 & 141,942 & 0.908 & 0.916 & \bf{0.924} & 0.731 & 0.570 & 0.897 \\
DBpedia work-genre               & 156,145 & 222,517 & 0.879 & 0.941 & 0.908 & 0.746 & 0.867 & \bf{0.966} \\
Epinions                         & 876,252 & 13,668,320 & 0.644 & \bf{0.690} & 0.546 & 0.501 & 0.061 & \bf{0.690} \\
French Wikipedia                 & 3,989,678 & 41,392,490 & 0.667 & 0.744 & 0.744 & 0.654 & 0.108 & \bf{0.803} \\
German Wikipedia                 & 3,357,353 & 51,830,110 & 0.673 & 0.699 & 0.699 & 0.651 & 0.156 & \bf{0.799} \\
Japanese Wikipedia               & 1,892,869 & 18,270,562 & 0.740 & 0.752 & 0.755 & 0.618 & 0.076 & \bf{0.776} \\
Jester                           & 25,038 & 616,912 & 0.575 & 0.571 & \bf{0.581} & 0.461 & 0.579 & 0.501 \\
MovieLens 100k                   & 2,625 & 100,000 & \bf{0.822} & 0.774 & 0.738 & 0.718 & 0.631 & 0.812 \\
MovieLens 10M                    & 136,700 & 10,000,054 & \bf{0.683} & 0.682 & 0.663 & 0.500 & 0.298 & 0.680 \\
MovieLens 1M                     & 9,746 & 1,000,209 & 0.640 & \bf{0.662} & 0.538 & 0.500 & 0.221 & \bf{0.662} \\
MovieLens tag-item               & 24,129 & 95,580 & 0.860 & 0.860 & 0.860 & 0.737 & \bf{0.865} & 0.863 \\
MovieLens user-item              & 11,610 & 95,580 & 0.755 & 0.741 & 0.728 & 0.659 & 0.674 & \bf{0.812} \\
MovieLens user-tag               & 20,537 & 95,580 & 0.782 & 0.798 & 0.798 & 0.672 & 0.663 & \bf{0.915} \\
Netflix                          & 497,959 & 100,480,507 & \bf{0.674} & 0.671 & 0.670 & 0.500 & 0.322 & 0.672 \\
Spanish Wikipedia                & 2,684,231 & 23,392,353 & 0.634 & 0.750 & 0.750 & 0.655 & 0.094 & \bf{0.799} \\
Wikipedia categories             & 2,036,440 & 3,795,796 & 0.591 & 0.659 & 0.663 & 0.500 & 0.589 & \bf{0.675} \\

    \hline
  \end{tabular}
  }
  \label{tab:results}
\end{table}

\subsection{Detecting Near-bipartite Networks}
Some networks are not bipartite, but nearly so.  An example would be a
network of ``fan'' relationships between persons where there are clear ``hubs'' and
``authorities'', i.e. popular persons and persons being fan of many
people.  While these networks are not strictly bipartite, they 
are mostly bipartite in a sense that has to be made precise.
Measures for the level of bipartivity exist in several
forms~\cite{b531,b456}, and
spectral transformations offer another method.  
Using the link prediction method described in Section~\ref{subsec:curve}, nearly
bipartite graphs can be recognized by the \emph{odd} shape of the
learned curve fitting function.  

\begin{figure}[t]
  \centering
  \subfigure[Advogato trust network]{\includegraphics[width=.49\textwidth]{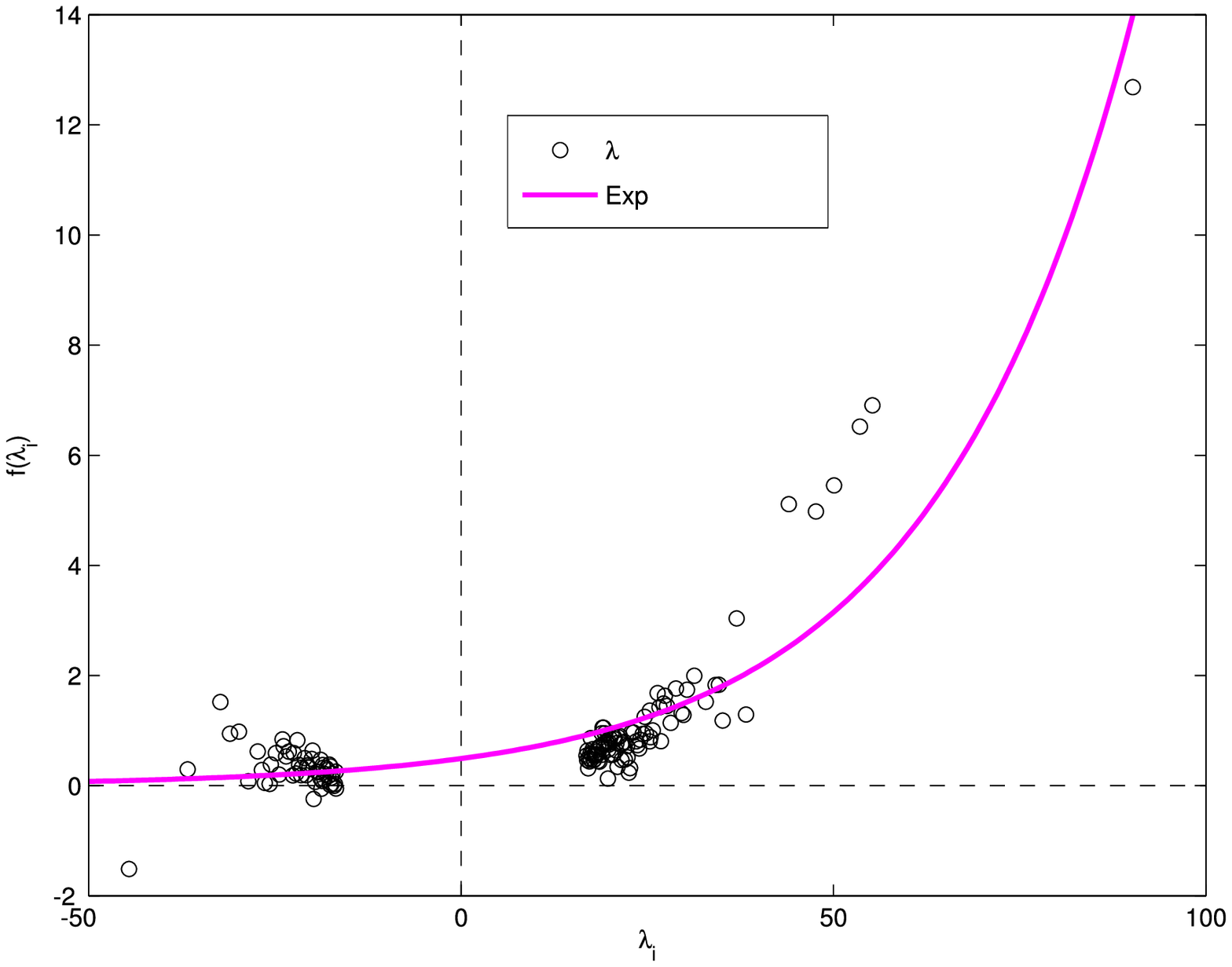}} 
  \subfigure[English Wikipedia hyperlinks]{\includegraphics[width=.49\textwidth]{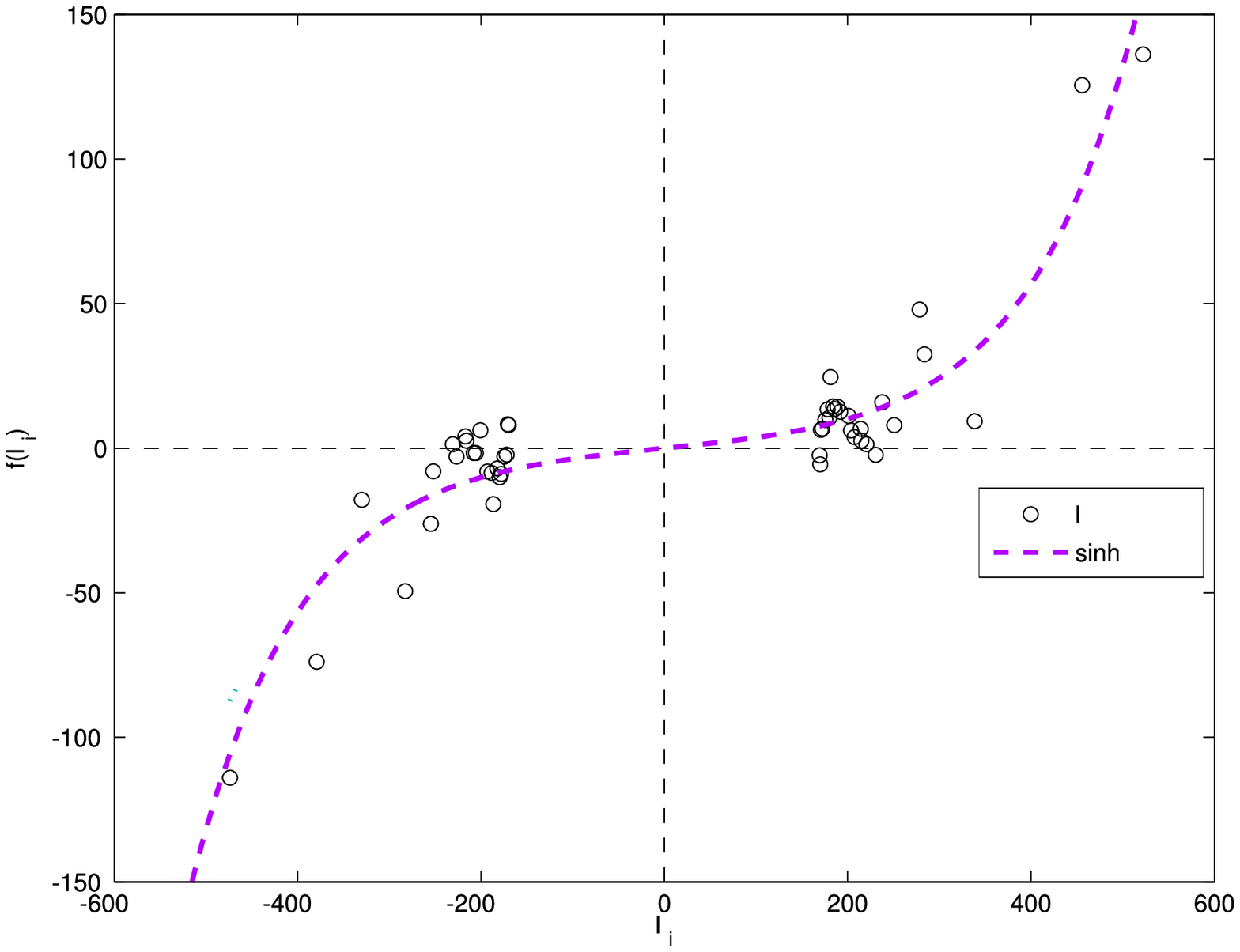}}
  \caption{
    Detecting near-bipartite and non-bipartite networks: If the
    hyperbolic sine fits, the network is nearly bipartite; if the
    exponential fits, the network is not nearly bipartite.  (a)~the
    Advogato trust network, (b)~the English Wikipedia hyperlink network.
    These graphs show the learned transformation of a graph's
    eigenvalues; see the text for a detailed description.
  }
  \label{fig:near-bipartite}
\end{figure}

Figure~\ref{fig:near-bipartite} shows the method applied to two
unipartite networks:  the Advogato trust network~\cite{b334} and the
hyperlink network in the English Wikipedia~\cite{download.wikimedia.org}.  The curves
indicate that the Advogato trust network is \emph{not} bipartite, while
the Wikipedia link network is nearly so.  

\section{Discussion}
\label{sec:discussion} 
While technically the link prediction problem in bipartite graphs is a
subproblem of the general link prediction problem, the special structure
of bipartite graphs makes common link prediction algorithms
ineffective.  In particular, all methods based on the triangle closing
model cannot work in the bipartite case.  Out of the simple local link
prediction methods, only the preferential attachment model can be used
in bipartite networks. 

Algebraic link prediction methods can be used instead, 
by restricting spectral transformations to odd functions,
leading to the matrix hyperbolic sine as a link prediction function, and an
odd variant of the von Neumann kernel. 
As in the unipartite case, no single link prediction method is best for
all datasets. 

\bibliographystyle{splncs}
\bibliography{kunegis}

\end{document}